\documentclass[journal]{IEEEtran}
\ifCLASSINFOpdf
 \else
 \fi
\usepackage{url} 
\usepackage{algorithm}
\usepackage{algpseudocode}

\usepackage{amsmath,amssymb}
\usepackage{multirow}
\usepackage{bbm}
\usepackage{graphicx}
\usepackage{float}
\usepackage{placeins}
\usepackage{booktabs}
\usepackage{orcidlink}
\begin{document}

\title{Walkable to Whom? Capturing Subjective Variability in Walkability Perception Using Multimodal Deep Learning}

\author{Moloud~Damandeh\,\orcidlink{0009-0001-4344-9858},~\IEEEmembership{Student Member,~IEEE}, Meead~Saberi\,\orcidlink{0000-0002-6526-239X}%
\thanks{This work was supported by the Australian Research Council under Grant DP220102382. M. Saberi was also supported by the Australian Research Council through Future Fellowship FT250100584.}%
\thanks{This work involved human subjects. Approval of all ethical procedures and protocols was granted by the University of New South Wales Human Research Ethics Committee under approval number 8361.}%
\thanks{Data and code is available at https://github.com/Moloudd/user-conditioned-walkability-assessment}%
\thanks{Conflict of Interest: M. Saberi is co-founder and CEO of footpath.ai, the imagery provider for this study, which had no role in study design or the decision to publish.}%
\thanks{The authors are with the School of Civil and Environmental Engineering, University of New South Wales (UNSW), Sydney, NSW, Australia, and the Research Centre for Integrated Transport Innovation (rCITI).}%
\thanks{Corresponding author: Meead~Saberi (e-mail: meead.saberi@unsw.edu.au).}%
}

\maketitle

\begin{abstract}
Visual perception of walkability varies substantially across individuals, reflecting differences in personal characteristics, experiences, and preferences. Existing studies, however, often reduce these diverse judgements to aggregated scores, implicitly assuming uniform perception, and commonly rely on vehicle-mounted street-view imagery that does not reflect the pedestrian's visual experience. This paper introduces a dataset of 29,870 walkability ratings from 1,196 respondents, linking sidewalk-view imagery across urban, suburban, and regional Australian environments with individual rater attributes, and proposes the first user-conditioned multimodal deep learning framework for walkability perception, fusing visual features with respondent-level representations. A viewpoint-comparison study shows that sidewalk-view images receive significantly higher walkability ratings than matched street-view images, indicating that imagery source is a substantive design decision in perception surveys. The user-conditioned model improves rank agreement with observed ratings by 65\% over an image-only baseline (quadratic weighted kappa 0.47 vs. 0.29), demonstrating that who is evaluating an environment carries predictive indication beyond image content alone. These findings support moving from aggregated, observer-independent walkability scores toward models that represent diverse users, enabling more inclusive assessment of pedestrian environments.
\end{abstract}
\IEEEpeerreviewmaketitle

\section{Introduction}
 \IEEEPARstart{W}{alkable} neighbourhoods support stronger social interaction and economic activity within communities, making walkability an important consideration in urban planning and design~\cite{duncan2011validation, zhou2019social}. Understanding how people perceive the walkability of their environment, and why those perceptions differ, is essential for designing streets that serve diverse communities~\cite{jehle2024does}.  While visual characteristics of the built environment play a key role in shaping perceived walkability~\cite{ito2024understanding, li2022measuring}, perception is not determined by environmental characteristics alone~\cite{jehle2024does, quintana2025global}. It is also shaped by user-specific factors, including demographic characteristics, geographic context, socio-cultural background, and mobility-related needs. As a result, the same environment may be perceived differently by different people \cite{jehle2024does, quintana2025global, dickinson2024geographic}, adding further complexity to the already difficult task of quantifying walkability~\cite{li2022measuring, dickinson2024geographic}. While prior work has made substantial progress in modeling urban perception from visual features ~\cite{dubey2016deep, costa2025cycling, blevcic2018towards}, predictive models that account for this subjective variability remain underexplored.
 
Another limitation of existing image-based methods for predicting walkability perception is their reliance on street-view imagery captured from vehicle-mounted cameras rather than from a pedestrian viewpoint. Such imagery can miss important details relevant to pedestrians. In contrast, sidewalk-view imagery provides a more pedestrian-centred perspective, reducing viewpoint bias and better capturing the elements that matter to active mobility users~\cite{ito2024translating}. However, whether this viewpoint mismatch affects perceived walkability ratings remains largely unexamined, even though recent studies have begun to establish protocols for image-based urban perception surveys \cite{gu2025designing}.   

Alongside standardized survey protocols, progress in this area also depends on the availability of open datasets that can support reproducible benchmarking~\cite{ito2024understanding}. Place Pulse 2.0 remains one of the most widely used large-scale public benchmarks for urban visual perception, providing high-quality perception annotations but limited respondent attributes~\cite{dubey2016deep}. More recently, SPECS has provided, to our knowledge, the only publicly available urban perception dataset that links perception ratings with richer respondent characteristics and analyses differences across demographic and personality groups~\cite{quintana2025global}. However, SPECS primarily supports group-level analyses of broader urban perception indicators. Publicly available datasets designed for individual-level, user-conditioned modeling of walkability perception remain scarce, particularly those pairing sidewalk-view imagery with walkability ratings and rater-specific attributes.

\begin{figure*}[!t]
\centering
\includegraphics[width=\textwidth]{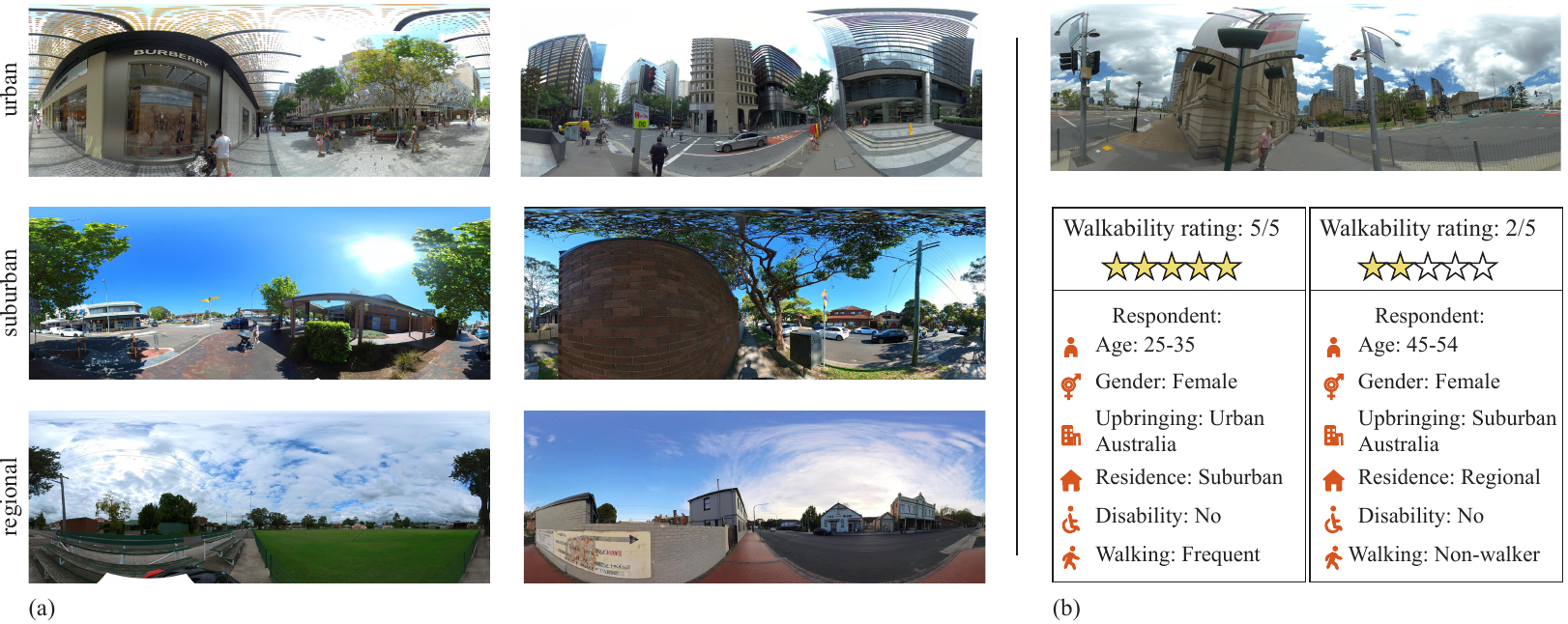}
\caption{Overview of the proposed dataset structure. (a) Representative sidewalk-view panoramic images sampled across urban, suburban, and regional environments. (b) Example image--respondent records showing how each sidewalk-view image is linked to  walkability ratings and respondent-level attributes.}
\label{fig:dataset}
\end{figure*}

This paper addresses these gaps through three main contributions. First, it introduces a new sidewalk-view walkability perception dataset linked with individual rater attributes. Second, it demonstrates that image viewpoint affects perceived walkability, showing that the choice of visual data source should be considered when designing surveys for walkability perception studies. Third, it proposes and evaluates a multimodal user-conditioned deep learning framework that fuses visual features with rater-level representations to model individual-level variability in walkability perception. By incorporating respondent attributes alongside image content, the framework demonstrates that subjective differences in perceived walkability can be learned and explained, providing a more personalized representation of visual walkability than image-only approaches. Beyond urban assessment and design, these user-conditioned predictions could also support personalized pedestrian routing, where predicted street-level walkability ratings enter route cost functions alongside travel time. This would enable different users to receive routes that better reflect their own perceptions and preferences, analogous to multi-criteria cycling routing approaches that incorporate subjective criteria such as comfort and safety~\cite{hrncir2017practical}.

 \section{RELATED WORK}
 \subsection{Visual Walkability Assessment}
 Visual walkability assessment has increasingly used street-level imagery and deep learning to capture physical features as experienced by pedestrians at street level~\cite{ito2024understanding}. A dominant methodological approach in image-based urban perception is crowdsourced pairwise comparison, where urban perception is formulated as a ranking problem. This paradigm was established using the Place Pulse 2.0 dataset, which contains 1.17 million pairwise comparisons across 56 cities, and a Siamese Convolutional Neural Network (CNN) trained to predict which of two streetscapes was perceived as safer, livelier, or more beautiful~\cite{dubey2016deep}. The same framework has since been adapted to walkability by collecting pairwise visual walkability ratings and training a multitask CNN to predict both overall walkability and contributing dimensions, including safety, comfort, and accessibility~\cite{li2022measuring}. A related comparison-based approach has also been applied to cycling safety perception, where a Siamese CNN was trained using pairwise comparisons, and deployed for city-wide assessment~\cite{costa2025cycling}. Other studies have used direct rating rather than pairwise comparison. For example, five-point Likert-scale ratings of Google Street View images (SVI) have been used to train a CNN to classify images into five ordinal walkability categories~\cite{blevcic2018towards}. However, these models characterise the visual environment independently of the observer, treating individual differences in perception as noise to be averaged out rather than as meaningful variation to be modeled. In this study, walkability ratings were collected using a five-point Likert scale. Unlike pairwise comparison methods, which yield image-level aggregate scores and require additional grouping to analyse demographic differences~\cite{quintana2025global}, direct ratings preserve respondent-level scores for each image. This is a structural requirement for the proposed user-conditioned modelling framework. The Likert scale also requires fewer ratings per image to obtain robust estimates~\cite{gu2025designing}, making it well-suited to the scale of this study.

\subsection{Observer-Level Variability in Urban Perception}
The concept of walkability often assumes that pedestrians form a homogeneous group, but empirical evidence increasingly challenges this assumption~\cite{dickinson2024geographic}. Cross-cultural studies show that although a broad understanding of walkability may be shared across groups, the specific built-environment attributes that define a walkable place vary systematically with sociocultural background and residential history ~\cite{dickinson2024geographic}. At a broader scale, a large-scale survey of participants across five countries found that streetscape perceptions vary significantly by sociodemographic attributes~\cite{quintana2025global}. The study further showed that models trained on aggregated responses can introduce systematic bias by masking this variation. Within pedestrian planning, studies similarly show that the attributes most important to walkability differ substantially across user groups with infrastructure quality and physical accessibility being especially salient for older pedestrians~\cite{jehle2024does}. Despite this evidence, current planning approaches that account for user differences typically rely on predefined group typologies derived from expert knowledge, rather than learning observer-specific differences from individual-level perceptual data ~\cite{jehle2024does}. At the same time, computational walkability models generally do not incorporate observer characteristics as inputs to prediction functions trained on empirical ratings. Consequently, the individual-level perceptual variability documented across these studies remains largely unrepresented in current predictive architectures.

\subsection{Personalized Subjective Image Assessment}
The task of predicting subjective image scores for individual observers, rather than a single aggregate score, has recently gained attention in the image aesthetics assessment literature ~\cite{yang2022personalized, maerten2025lapis}. A key distinction in this field is between generic image aesthetic assessment, which predicts a mean opinion score across all raters, and personalized image aesthetic assessment, which predicts observer-specific scores~\cite{maerten2025lapis}. Early personalized image aesthetic assessment methods primarily used image-level attributes to adjust generic predictions toward individual preferences~\cite{ren2017personalized}. More recent work incorporates personal attributes such as demographics and personality traits into prediction models. The PARA dataset, for example, includes image ratings from 438 subjects along with attributes such as age, gender, education, and Big Five personality scores \cite{yang2022personalized}. Models trained on PARA show improved performance when conditioned on subject information. Similarly, the LAPIS dataset provides rich annotator attributes for artistic images, and ablation studies show that removing certain attributes degrades prediction performance \cite{maerten2025lapis}, confirming that observer characteristics carry predictive signal beyond image content alone. In the transportation planning domain, existing walkability and urban perception models remain closer to the generic formulation. They typically produce a single aggregate score per image or location, treating all observers as equivalent. User-conditioned prediction, in which explicit observer attributes are jointly encoded with image features, has therefore not yet been applied to walkability or pedestrian environment perception. Learning individual user preferences has also been recognised in applications beyond visual perception, including personalised multimodal route planning systems, where user-specific preferences are learned to improve route choice modelling~\cite{arentze2013adaptive}. More broadly, multimodal fusion of heterogeneous data sources has been identified as an important direction within the intelligent transportation systems (ITS) deep learning literature \cite{veres2020deep}. However, prior applications have predominantly focused on transportation forecasting, perception, and control. The present study extends multimodal learning to pedestrian walkability perception by jointly encoding sidewalk-view imagery and respondent-level characteristics.

\subsection{Sidewalk-View Imagery and View-Point Bias}
Street-view imagery, also known as SVI, has become an important data source in urban research because it enables scalable assessment of street-level built-environment characteristics that are difficult to capture using conventional GIS-based measures~\cite{ito2024translating,biljecki2021street}. However, most existing SVI-based studies rely on imagery collected from vehicle-mounted cameras~\cite{qin2024crowd}. This introduces a potential viewpoint bias, because vehicle-based imagery represents the street from the carriageway rather than from the pedestrian path. Recent studies have begun to quantify this limitation by comparing semantic information extracted from street-view and sidewalk-view imagery. These studies report weak or inconsistent correspondence between the two viewpoints, suggesting that the visual features visible from the vehicle lane can differ substantially from those encountered on the sidewalk~\cite{ito2024translating, rui2023measuring}. Despite growing evidence that viewpoint affects the visual representation of the street environment, existing image-based walkability studies have largely relied on vehicle-based SVI or static image views. More importantly, little is known about whether people evaluate the same location differently when it is presented from a street-view versus a sidewalk-view perspective. This leaves unresolved whether viewpoint differences translate into differences in perceived walkability for the same location.
 
\section{Methodology}
\subsection{Dataset}
\subsubsection{Image Selection}
Sidewalk-view imagery was sourced from footpath.ai\footnote{\url{https://footpath.ai}}, a large-scale provider of high-resolution 360-degree street-level imagery. The source dataset includes image-level sidewalk and pedestrian infrastructure attributes relevant to walkability assessment, extracted using  AI-based semantic segmentation and object detection methods. These attributes include greenery, sky, building coverage, shaded areas, sidewalk width, benches, garbage bins,  marked crossings, and accessibility curb cuts. They are key dimensions of the physical pedestrian environment \cite{ewing2009measuring} and were used to guide a representative image sampling strategy.

To capture variability in walkability perception across settlement types, we selected nine Australian locations spanning urban, suburban, and regional contexts.  Representative examples of sidewalk-view images across settlement types are shown in Fig.~\ref{fig:dataset}(a). This sampling design extends beyond the city-centre focus common in existing visual perception datasets \cite{dubey2016deep, quintana2025global}, none of which include suburban or regional settlement types. 

From the image pool across the selected settlement types, a representative subset was constructed using Threshold-Constrained Stratified Sampling (TCSS). Unlike semantic clustering approaches that group images by dominant visual composition~\cite{gu2025designing}, TCSS stratifies images based on fine-grained combinations of walkability-relevant attributes. This ensures that the sample preserves the real-world distribution of pedestrian infrastructure conditions, including sidewalk width, tree presence, and crossing availability, rather than generic visual diversity. The procedure first partitions the source pool into strata defined by unique combinations of the image-level attributes described above. Low-frequency strata are then aggregated to reduce excessive fragmentation, and a binary search procedure is used to identify the minimum sample size satisfying a predefined distributional constraint, with $\delta = 0.05$. Under this constraint, the marginal distribution of each walkability-relevant attribute in the final sample deviates from the corresponding distribution in the source pool by no more than 5\%. This tolerance was selected to balance sample compactness and distributional fidelity. The procedure yielded a final dataset of 974 images reflecting the diversity of pedestrian infrastructure conditions across the selected locations. 
\subsubsection{Online Survey}

An online survey was designed and conducted to collect human perceptions of walkability for sidewalk-view images. The survey consisted of two components. First, participants provided self-reported information, including age, gender, current place of residence, childhood residential environment, walking frequency, country of upbringing, and disability or health condition affecting walking. Second, each participant rated a block of 25 sidewalk-view images, pre-assigned to ensure a representative mix of urban, suburban, and regional environments. Images were displayed as interactive 360-degree panoramic 
views, allowing respondents to rotate and examine the 
sidewalk environment from multiple perspectives before 
submitting a rating on a five-point Likert scale in response to the prompt, "On a scale of 1 (poor walkability) to 5 (excellent walkability), how walkable does this environment appear to you?" 

To support consistent interpretation while allowing subjective judgement, an information icon was displayed alongside each image with the following definition: “Walkability refers to how suitable an area is for walking based on its overall environment and conditions. People may perceive walkability differently depending on their own experiences and expectations.” This definition was intentionally broad, acknowledging that individuals may weigh environmental attributes differently depending on their residential background, walking habits, and mobility-related needs. This design is consistent with the objective of modeling individual-level variability in walkability perception rather than imposing a fixed expert-defined criterion. Fig.~\ref{fig:dataset}(b) illustrates the resulting record structure, in which each sidewalk-view image is linked to an individual respondent's attributes and their walkability rating.

To ensure settlement-type diversity within each participant's rating block, images were assigned using a round-robin procedure. The image pool was first shuffled and grouped by settlement type and sampling stratum. Images were then sequentially allocated across rating blocks, ensuring that each block included urban, suburban, and regional images while maintaining diversity across the sampling strata.

Participants were recruited primarily through Pureprofile\footnote{\url{https://www.pureprofile.com}}, a commercial online research panel provider, and were compensated upon completing the survey. To broaden participation, additional voluntary responses were collected through professional and social networking channels. Recruitment aimed to obtain a geographically and socio-demographically diverse respondent pool across urban, suburban, and regional contexts, corresponding to the settlement-type categories represented in the image dataset.

Respondent reliability was assessed using multiple quality-control criteria. First, two randomly selected images were repeated within each survey session as consistency checks. Respondents were excluded if their ratings for both repeated images were inconsistent with their original ratings. Second, speeders were identified based on completion times below a minimum plausible threshold. Third, straightliners were identified as respondents who submitted the same rating to all image-rating items. Additional survey controls were implemented to reduce automated and duplicate responses \cite{qualtrics2024response}. All quality-control filters were applied before analysis. The final dataset comprised 1,196 valid participants. The average survey completion time was approximately 10 minutes.

\subsection{Viewpoint Comparison Survey}

\begin{figure*}[!t]
\centering
\includegraphics[width=\textwidth]{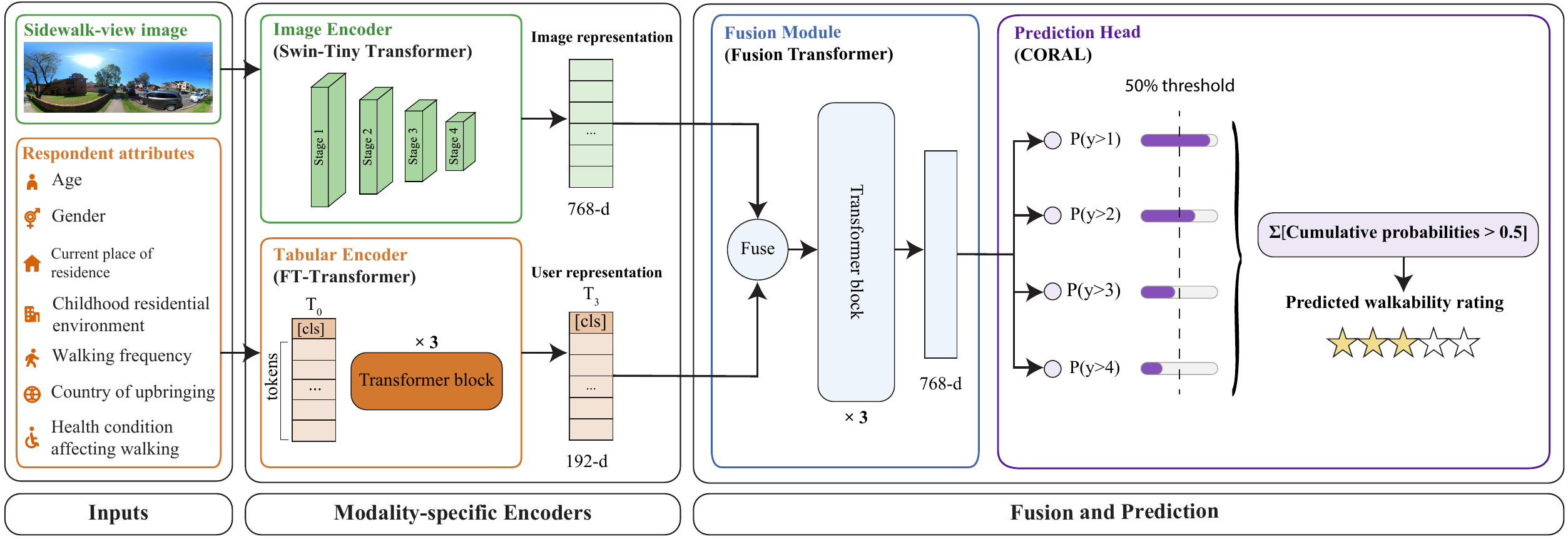}
\caption{Overview of the proposed user-conditioned multimodal framework for walkability perception prediction. The study evaluates alternative image backbones, tabular encoders, fusion mechanisms, and loss functions; the configuration shown corresponds to the best-performing model.}
\label{fig:pipeline}
\end{figure*}
To validate the use of sidewalk-view imagery as a more appropriate data source for walkability perception modeling, a comparison study was conducted using matched sidewalk-view and street-view image pairs from 100 locations. For each location, a sidewalk-view image from the proposed dataset was paired with a corresponding Google SVI. The selected locations were distributed evenly across the three settlement types represented in the main survey, ensuring coverage of urban, suburban, and regional environments. Each participant rated 20 images, consisting of one sidewalk-view and one street-view image from each of 10 randomly assigned locations, using the same five-point Likert scale as the main survey. Participants were not informed of the viewpoint distinction in order to reduce priming effects. Because ratings were paired by location, measured on an ordinal scale, and not normally distributed, a Wilcoxon signed-rank test was used to assess whether viewpoint systematically influenced perceptions of walkability.

\subsection{User-Conditioned Multimodal Deep Learning Framework for Walkability Perception}
\subsubsection{Problem Formulation}
Let $I_i$ denote a sidewalk-view image and let 
$\mathbf{x}_j \in \mathbb{R}^{d}$ denote a respondent-level attribute vector 
associated with participant $j$. The attribute vector encodes individual 
characteristics that may influence walkability perception, including demographic 
information, residential background, walking habits, and mobility-related 
conditions. For each image--respondent pair $(I_i,\mathbf{x}_j)$, the observed 
target is a perceived walkability rating $y_{ij} \in \{1,2,3,4,5\}$ on a 
five-point ordinal Likert scale. 

Motivated by evidence that walkability perception varies systematically with 
individual characteristics~\cite{quintana2025global}, and by the demonstrated effectiveness of conditioning visual assessments on observer-level attributes to model individual differences in subjective perception~\cite{yang2022personalized}, the objective is to 
learn a user-conditioned prediction function:
\begin{equation}
    \hat{y}_{ij} = f(I_i,\mathbf{x}_j),
\end{equation}
where $\hat{y}_{ij}$ represents the predicted walkability rating assigned by 
respondent $j$ to sidewalk image $I_i$. This formulation differs from 
conventional image-based approaches that assign a single aggregate score to each 
location independent of the observer ~\cite{dubey2016deep,blevcic2018towards}. Instead, it requires the model to jointly 
encode visual environmental cues and respondent-level profile information, 
allowing perceived walkability to be modeled as an interaction between the 
sidewalk environment and the individual evaluating it.

\subsubsection{Multimodal Representation Learning}The formulation requires learning from two heterogeneous sources of information that differ in both data format and semantic role. The image describes the pedestrian environment being rated, whereas the respondent attributes describe the individual performing the rating. Multimodal learning from heterogeneous sources, particularly the combination of visual and structured tabular data, commonly relies on modality-specific encoders that project each input into a learned representation space before fusion~\cite{ebrahimi2023lanistr}.

Fig.~\ref{fig:pipeline} illustrates the proposed framework. The image encoder $g_I(\cdot)$ maps each sidewalk-view image into a latent visual
representation:
\begin{equation}
    \mathbf{z}^{I}_{i} = g_I(I_i),
\end{equation}
capturing pedestrian-environment cues relevant to perceived walkability. In
parallel, the respondent encoder $g_X(\cdot)$ maps the attribute vector into a
latent user representation:
\begin{equation}
    \mathbf{z}^{X}_{j} = g_X(\mathbf{x}_j),
\end{equation}
encoding individual-level characteristics that condition the perception of
the same environment. 

The two representations are combined through a learned fusion module:
\begin{equation}
    \mathbf{z}_{ij} =
    h\left(\mathbf{z}^{I}_{i},\,\mathbf{z}^{X}_{j}\right),
\end{equation}
where $h(\cdot,\cdot)$ is a learned function that combines the two modality-specific representations, enabling the model to learn interactions between visual environmental features and respondent characteristics in a shared representation space before prediction.

For the image encoder $g_I(\cdot)$, we adopt swin-tiny~\cite{liu2021swin}, a hierarchical vision Transformer based on shifted-window self-attention, pretrained on ImageNet-1K. Each 360° image is stored as an 2:1 equirectangular panorama, and processed according to the pretrained Swin-Tiny input configuration: the shorter side is resized to 224 pixels, after which a centred 224×224 crop is used as the network input, retaining the central 180° horizontal field of view. This follows prior work applying standard 2D vision architectures directly to panoramic imagery without geometric correction \cite{li2022measuring}. For the respondent encoder $g_X(\cdot)$, we employ FT-Transformer~\cite{gorishniy2021revisiting}, a transformer-based model designed for tabular data. The fusion module $h(\cdot, \cdot)$ is implemented as a Transformer-based module~\cite{vaswani2017attention} projecting each modality's representation to a common 768-dimensional space, and processing the result through 3 Transformer blocks to obtain a 768-dimensional joint embedding. The justification for these specific choices is provided through ablation experiments in Section ~\ref{subsec:ablation}.

\subsubsection{Ordinal Prediction} The target variable $y_{ij}$ is an ordered categorical rating rather than a nominal class. Treating this task as standard multiclass classification ignores the ordinal structure of the scale and treats all incorrect predictions as equally distinct, regardless of their distance on the rating scale. This is undesirable for perceived walkability ratings, where adjacent categories may reflect minor differences in judgment, while larger ordinal deviations indicate more substantial disagreement between predicted and observed perception.

To preserve the ordered structure of the response variable, the framework 
follows the ordinal regression formulation of CORAL~\cite{CAO2020325}, which decomposes 
a $K$-class ordinal prediction problem into $K-1$ binary threshold tasks while 
enforcing rank-consistent predictions. Rank consistency ensures that if the model predicts the rating exceeds threshold $k$, it must also predict that it exceeds all lower thresholds $k'< k$. For the five-point walkability scale used in this study, $K=5$, and each observed rating $y_{ij}$ is transformed into 
a set of binary ordinal labels:
\begin{equation}
    r_{ij,k} = \mathbbm{1}(y_{ij} > k),
    \quad k \in \{1,2,3,4\}.
\end{equation}
The model estimates the probability that the perceived walkability rating 
exceeds each ordinal threshold:
\begin{equation}
    p_{ij,k} = P(y_{ij} > k \mid I_i, \mathbf{x}_j).
\end{equation}

The final predicted rating is then recovered by summing the predicted threshold outcomes:
\begin{equation}
    \hat{y}_{ij} =
    1 + \sum_{k=1}^{K-1} \mathbbm{1}(p_{ij,k} > 0.5).
\end{equation}

The framework is trained end-to-end by minimising the sum of binary
cross-entropy losses across the $K-1$ ordinal thresholds:
\begin{equation}
\mathcal{L}_{\text{ord}} =
-\sum_{i,j}\sum_{k=1}^{K-1}
\left[
r_{ij,k}\log p_{ij,k}
+
(1-r_{ij,k})\log(1-p_{ij,k})
\right].
\end{equation}

\section{Results}
\label{sec:results}
\subsection{Analysis of Dataset}
The final dataset contains 29,870 valid image--respondent ratings for 974
sidewalk-view images. Each image was rated by 25 to 35 distinct respondents,
with an average of 31 ratings per image. This exceeds the minimum number of ratings per image recommended for reliable Likert-scale image-based perception surveys~\cite{gu2025designing}.  The respondent pool includes variation in age, gender, current residential context, childhood residential context, walking frequency, country of upbringing, and disability or health condition affecting walking. This diversity supports the modeling of individual-level variation in walkability perception, while recognising that some groups are represented less frequently than others.

\subsubsection{Rating distribution}
\begin{figure}[t]
\centering
\includegraphics[width=\columnwidth]{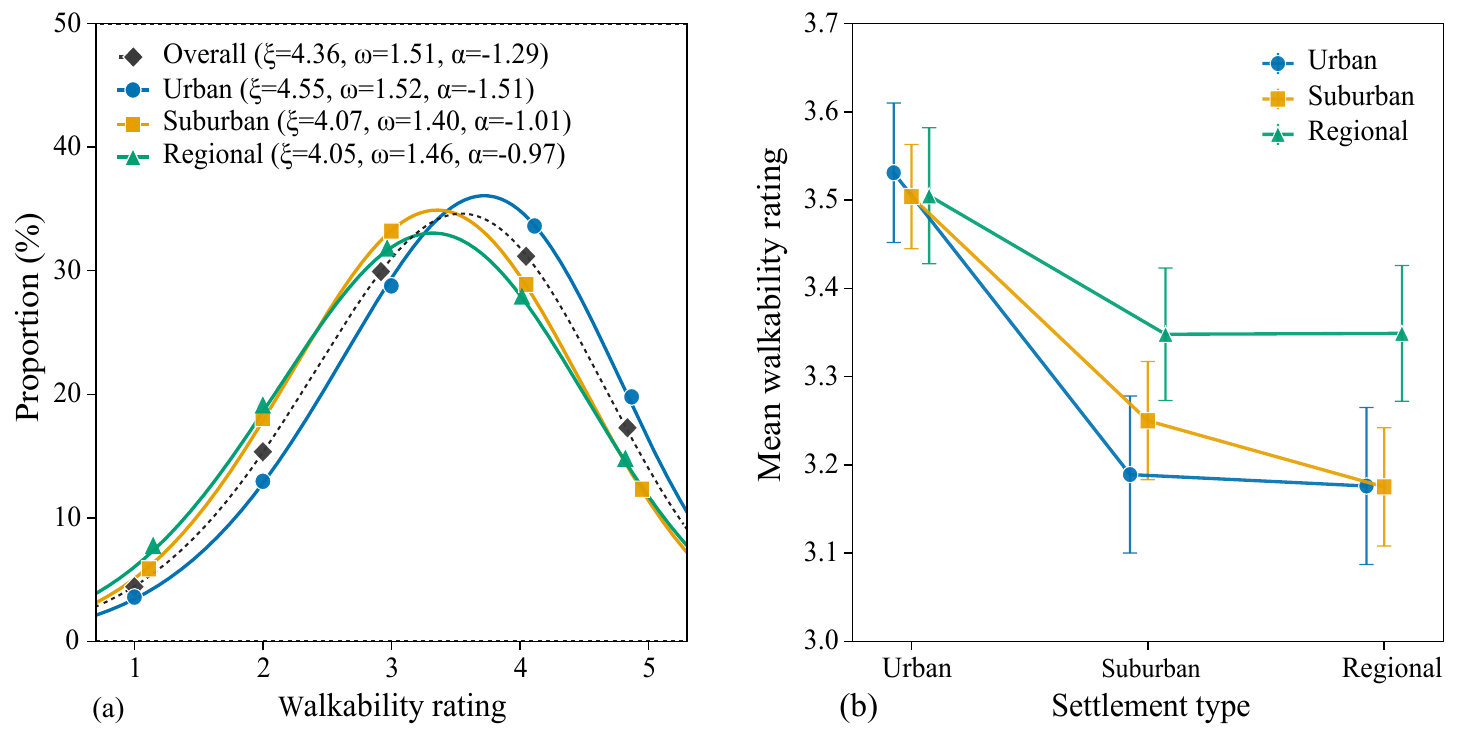}
\caption{Perceived walkability ratings by image settlement type.
(a) Distribution of ratings overall and for urban, suburban, and regional
imagery. Symbols denote the observed proportion of responses at each rating;
Solid (dashed for Overall) curves are skew-normal distributions fitted to each
group using a common parametric family, parameterized by location ($\xi$),
scale ($\omega$), and shape ($\alpha$). (b) Interaction between image settlement
type and respondent residential context; points represent respondent-level mean
walkability ratings and error bars indicate 95\% confidence intervals.}
\label{fig:dist}
\end{figure}

Figure~\ref{fig:dist}(a) presents the distribution of perceived walkability ratings overall and by settlement type, with a skew-normal distribution fitted to each group using a common parametric family to enable direct comparison. Ratings were concentrated in the middle categories, with scores 3 and 4 accounting for 61.1\% of all responses, while score 1 represented only 5.8\% of ratings. This distribution reflects a moderate class imbalance and is consistent with patterns observed in subjective image assessment datasets \cite{maerten2025lapis}, partially attributable to respondents' tendency to avoid the extremes of rating scales \cite{rezende2022rating}. The fitted distributions captured this structure well. All four groups exhibited a negative shape parameter ($\alpha < 0$), indicating left-skewed distributions in which responses concentrated toward the middle-to-moderately-high part of the scale with only a thin tail extending toward low ratings.

Differences were also observed across settlement types. Urban images showed a higher concentration of ratings in the upper categories, peaking at rating 4, with only 4.4\% assigned to category 1. In contrast, suburban and regional images peaked at rating 3, and contained a higher proportion of lower ratings. These differences were reflected in the fitted parameters. The urban distribution had the highest location ($\xi$) and the strongest negative skew, whereas the suburban and regional distributions were near-identical to one another and shifted toward lower ratings. Notably, the scale parameter ($\omega$) was comparable across all groups, indicating that the distributions differed primarily in central location rather than in dispersion or overall shape. This trend suggests that perceived walkability varies systematically across settlement contexts, potentially reflecting differences in pedestrian infrastructure quality and sidewalk conditions across urban, suburban, and regional environments.

\subsubsection{Inter-Rater Variability}Inter-rater agreement was quantified using pairwise Cohen's quadratic weighted kappa. This metric is appropriate for ordinal Likert-scale ratings because it accounts for chance agreement while assigning larger penalties to disagreements that are farther apart on the rating scale ~\cite{gu2025designing}. Across all eligible respondent pairs (pairs who both rated at least five common images), the mean quadratic weighted kappa was 0.17, indicating limited agreement among respondents when evaluating the same sidewalk environments. This value is lower than kappa values reported in studies using more homogeneous participant pools and smaller image sets~\cite{gu2025designing},  reflecting the greater socio-demographic diversity of respondents and environmental variability of images in the present study.

To assess whether this agreement exceeded chance levels, we compared the observed kappa against a permutation-based null distribution obtained by randomly shuffling image assignments across ratings. The observed agreement was higher than expected under the null model (mean null $\kappa \approx 0.000$, $SD = 0.026$, $p < 0.001$), suggesting that respondents share some common visual interpretation of walkability while still exhibiting substantial individual-level variation. This finding supports the use of a user-conditioned modeling framework rather than relying solely on aggregate image-level scores.

\subsubsection{Respondent–Environment Interactions in Walkability Perception}We examined the relationship between respondent-level attributes and perceived walkability ratings using Spearman correlation for ordinal attributes and Kruskal--Wallis tests for categorical attributes. No strong main effects were observed for age, gender, residential context, or disability affecting walking. Walking frequency showed statistically significant but negligible association with mean rating ($r=0.06$, $p=0.034$), suggesting that respondent attributes do not uniformly shift overall rating levels.

In contrast, significant interaction effects were observed between image settlement type and several respondent attributes, based on mixed-effects likelihood-ratio tests with random intercepts for respondent and image. Image settlement type interacted significantly with current residential context ($\text{LR } \chi^2(4) = 58.58$, $p < 0.001$), age ($\text{LR } \chi^2(10) = 51.83$, 
$p < 0.001$), and walking frequency ($\text{LR } \chi^2(8) = 63.94$, $p < 0.001$). As shown in Fig.~\ref{fig:dist}(b), respondents from different residential contexts rated urban images similarly, but diverged for suburban and regional images, with urban-dwelling respondents assigning lower ratings to non-urban environments. Comparable patterns were observed for age and walking frequency, indicating that respondent attributes affect how individuals differentiate between sidewalk environments rather than simply shifting their overall rating tendency.
These patterns are consistent with prior work on subjective image assessment datasets \cite{maerten2025lapis}, where observer-level attributes may influence ratings through interactions with image content rather than through uniform shifts in mean scores. The observed interactions provide empirical support for conditioning walkability prediction on both sidewalk-view image content and respondent-level characteristics.

\begin{figure}
\centering
\includegraphics[width=\columnwidth]{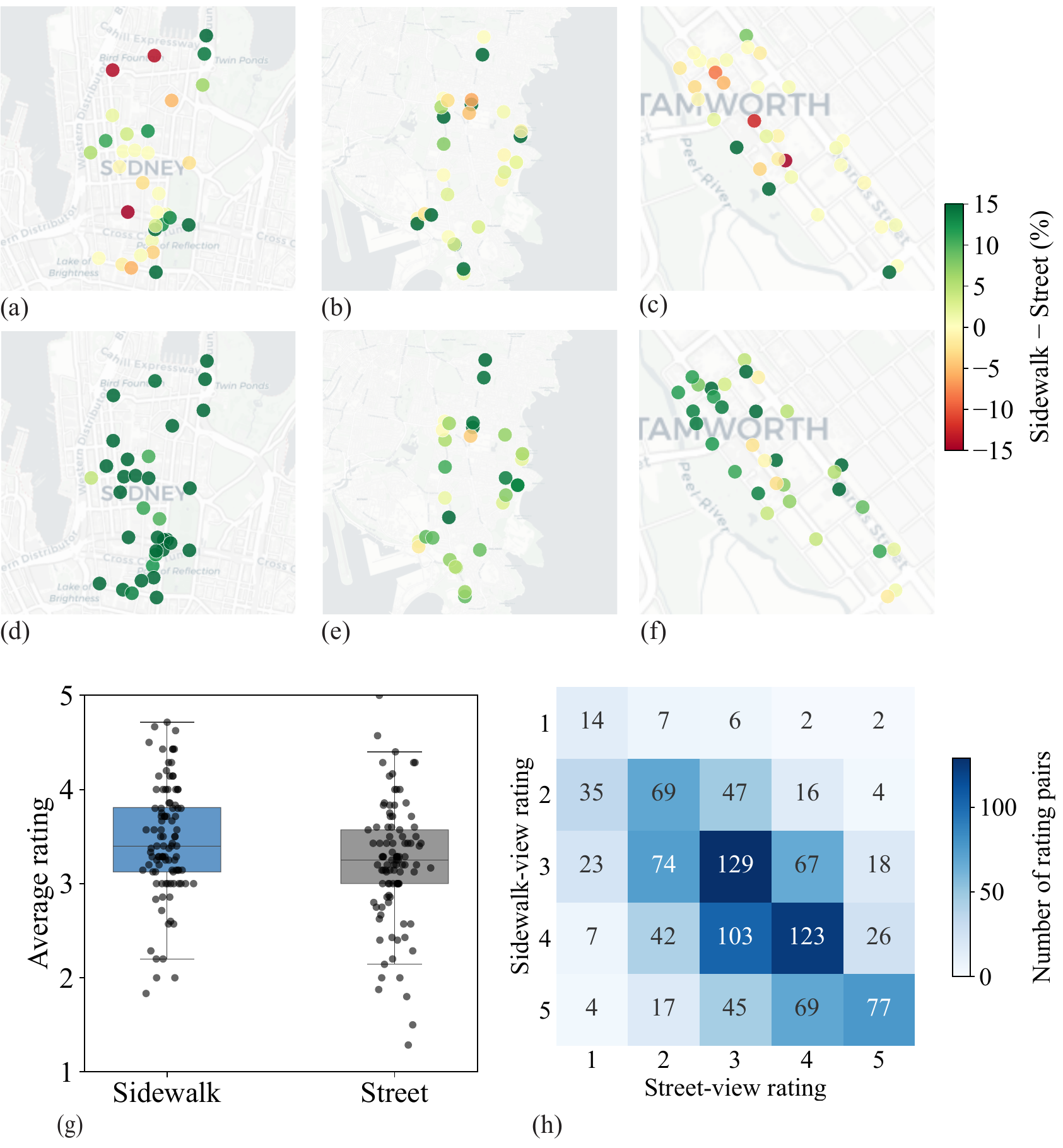}
\caption{Viewpoint comparison across 100 matched locations. (a--c) Sidewalk and (d--f) tree pixel coverage difference (sidewalk minus street, \%) across Sydney CBD, Maroubra, and Tamworth, shown as representative examples of consistent and heterogeneous viewpoint effects respectively. (g) Distribution of mean walkability ratings per image. (h) Paired rating combination frequencies across 1,076 individual rating pairs.}
\label{fig:viewpoint}
\end{figure}

\subsection{View point comparative analysis}
Figure ~\ref{fig:viewpoint}(a)--(f) presents representative spatial distributions of differences in sidewalk and tree pixel coverage between matched sidewalk-view and street-view images across the three study areas. Sidewalk coverage was generally higher in sidewalk-view imagery, particularly in urban and regional areas, while a more spatially variable pattern was observed in the suburban areas. In contrast, tree coverage did not exhibit a consistent directional pattern across settlement types. These results indicate that viewpoint differences do not uniformly increase or decrease the visibility of all visual features; rather, their effects vary with the local street environment, consistent with evidence that perspective differences have varying geographic effects on urban perceptions \cite{rui2023measuring}.

Figure~\ref{fig:viewpoint}(g) presents the distribution of mean walkability ratings per image for each viewpoint. On average, sidewalk-view images received higher walkability ratings than their matched street-view counterparts (mean = 3.48 vs. 3.16). A Wilcoxon signed-rank test conducted on location-aggregated median pairs confirmed that this difference was statistically significant ($W=207.5$, $Z=4.385$, $p<0.001$), with a large effect size (rank-biserial r=0.61, location-level Mdn = 3.50 vs. 3.00). Among the 100 matched locations, 52 exhibited non-zero median differences, while 48 had identical median ratings across the two viewpoints. At the observation level, 40.1\% of individual rating pairs were higher for the sidewalk-view image, 41.4\% were identical, and 18.6\% were higher for the street-view image, as shown in Fig.~\ref{fig:viewpoint}(h). When ratings differed, the most common shift was a one-point increase in favour of the sidewalk-view perspective, suggesting that the effect is consistent in direction but modest in magnitude. The higher ratings observed for sidewalk-view imagery, together with the feature-level differences identified in the matched image pairs, suggest that pedestrian-level imagery better captures aspects of the walking environment that are relevant to perceived walkability. These findings support the use of sidewalk-view imagery as the primary visual input for modeling walkability perception.
\subsection{Model results}
\subsubsection{Experimental Setup}
The dataset was divided into training, validation, and test subsets using a 70/10/20 split. Stratified sampling was applied based on walkability score and settlement type to preserve comparable distributions across the three subsets. This ensured that the validation and test sets closely followed the distribution of the training set with respect to both perceived walkability ratings and settlement categories. A representative test set is essential for obtaining a reliable estimate of model performance. Since the dataset is imbalanced toward middle-range walkability scores, a model biased toward predicting average ratings may still achieve seemingly reasonable performance. Therefore, an unrepresentative test set could lead to misleading conclusions about the model’s predictive ability. By applying stratified sampling, we ensured that the test set covers the full range of walkability scores and avoids disproportionate representation of images that may be easier for the model to predict.

The model was evaluated under a rating-completion protocol: raters and images could recur across the training and test partitions, but no individual image--respondent rating was shared between them. This setting was used to assess whether conditioning on respondent attributes improves the prediction of individual walkability judgments over a non-user-conditioned, image-only model, while keeping the image distribution comparable across partitions. Because the model was conditioned on respondent attributes rather than respondent identity, individual raters could not be directly memorized; the observed improvement was therefore interpreted as evidence of learnable structure in how respondent characteristics shape walkability perception. Accordingly, the results were framed as evidence for user-conditioned preference modelling within the sampled image pool, with broader generalisation left to future work with larger-scale data.

All models were implemented within the AutoGluon-Multimodal (AutoMM) framework~\cite{tang2024autogluon}, extended to support ordinal prediction. All models were trained using AdamW with a learning rate of $1 \times 10^{-4}$, weight decay of $1 \times 10^{-3}$, layer-wise learning-rate decay with a factor of 0.9, gradient-norm clipping at 1.0, and a cosine decay schedule with linear warm-up over the first 10\% of training steps~\cite{tang2024bag}. Training was run for up to 50 epochs with early stopping using a patience of 10 validation checks based on quadratic weighted
kappa (QWK). To account for training variability, all models were trained over five independent runs with different random seeds, and results are reported as mean ± standard deviation. All runs used mixed-precision (bf16) training with an effective batch size of 128.
\subsubsection{Evaluation Criteria}Because the prediction target was ordinal, evaluation metrics needed to reflect the ordered structure of the rating scale. Exact-match accuracy (ACC) was reported as a secondary indicator, as it treated all misclassifications equally, regardless of their distance from the true rating. We therefore reported mean absolute error (MAE) to quantify the average rating-level deviation between predicted and observed scores, and within-one accuracy to measure the proportion of predictions that fell within one rating level of the ground truth. However, within-one accuracy could be overly permissive for imbalanced five-point rating distributions, particularly when observations were concentrated in the middle categories. A model biased toward central ratings could still achieve high within-one accuracy without capturing the full ordinal structure of the data. For this reason, QWK was used as the primary model agreement metric between predicted and observed walkability ratings. This use follows common practice in ordinal prediction tasks, where model outputs are evaluated against reference human-assigned scores \cite{taghipour2016neural, sahlsten2019deep}.

\subsubsection{Effect of User Conditioning}
\begin{figure}[!t]
\centering
\includegraphics[width=\columnwidth]{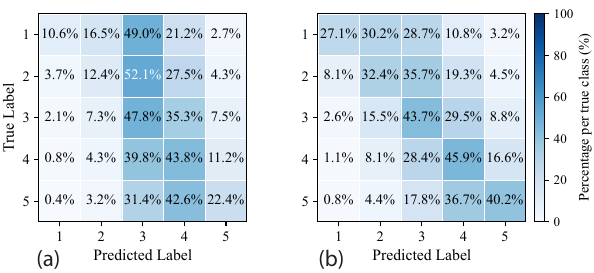}
\caption{Row-normalized test confusion matrices averaged over seeds for (a) the image-only baseline and (b) the proposed user-conditioned image-tabular model.}
\label{fig:confusion}
\end{figure}

Table \ref{tab:main_comparison} and Figure \ref{fig:confusion} present the main comparison between the image-only baseline and the proposed user-conditioned walkability perception framework. Incorporating respondent attributes yields substantial improvements across all metrics. Compared with the image-only baseline, QWK increases from 0.285$\pm$0.023 to 0.469$\pm$0.026, corresponding to an approximately 65\% relative improvement in rank agreement. MAE decreases from 0.885$\pm$0.014 to 0.785$\pm$0.029, while accuracy increases from 0.344$\pm$0.005  to 0.410$\pm$0.018. The improvement in QWK is particularly meaningful because this metric penalizes large ordinal errors quadratically, indicating that user conditioning not only improves classification performance but also reduces the severity of ordinal prediction errors.

Per-class results in table~\ref{tab:perclass_test_meanstd} reveal that the aggregate improvement is driven primarily by gains at the rating extremes. For classes 1 and 2, representing environments perceived as least walkable, recall increases from $0.106 \pm 0.035$ to $0.271 \pm 0.065$ and from $0.124 \pm 0.085$  to $0.324 \pm 0.019$ , respectively, with corresponding improvements in F1. Class 5 shows a similar pattern, with recall increasing from $0.224 \pm 0.030$ to $0.402 \pm 0.016$ and F1 from $0.278 \pm 0.017$ to $0.422 \pm 0.022$. These gains suggest that respondent attributes contribute discriminative information, particularly at the extremes of the walkability scale, where the image-only model shows the weakest recall. The exception is class 3, where the image-only model achieves marginally higher recall ($0.478 \pm ±0.097$ vs. $0.437 \pm ±0.012$); given that mid-range classes account for 61.1\% of ratings, this likely reflects a central tendency bias rather than stronger visual discriminability.  This interpretation is supported by the confusion matrices in Fig.~\ref{fig:confusion}, where the image-only model concentrates predictions toward classes 3 and 4, whereas the user-conditioned model produces a more distributed prediction profile across the rating scale. Overall, these results suggest that respondent-level attributes provide meaningful information for modelling individual differences in perceived walkability and help mitigate the central tendency bias observed when predictions rely only on visual features.

\begin{table}[!t]
\centering
\caption{Test-set comparison between the image-only baseline and the proposed user-conditioned model. W-1 denotes within-one accuracy.}
\label{tab:main_comparison}
\footnotesize
\renewcommand{\arraystretch}{1.1}

\begin{tabular*}{\columnwidth}{@{\extracolsep{\fill}}lcc@{}}
\hline
Metric & Image-only & User-conditioned \\
\hline
Input Mode(s) & Image & Image + Tabular \\
\hline
QWK$\uparrow$
& 0.285$\pm$0.023
& \textbf{0.469$\pm$0.026} \\
MAE$\downarrow$
& 0.885$\pm$0.014
& \textbf{0.785$\pm$0.029} \\
Acc.$\uparrow$
& 0.344$\pm$0.005
& \textbf{0.410$\pm$0.018} \\
W-1$\uparrow$
& 0.802$\pm$0.007
& \textbf{0.835$\pm$0.011} \\
\hline
\end{tabular*}
\end{table}

\begin{table}[!t]
\centering
\caption{Per-class test-set performance.}
\label{tab:perclass_test_meanstd}
\renewcommand{\arraystretch}{1.2}
\resizebox{\columnwidth}{!}{%
\begin{tabular}{clcc}
\toprule
Class & Metric & Image-only & User--conditioned \\
\midrule
\multirow{3}{*}{1}
 & Rec.$\uparrow$      & $0.106 \pm 0.035$ & \textbf{\boldmath$0.271 \pm 0.065$} \\
 & F1$\uparrow$        & $0.152 \pm 0.037$ & \textbf{\boldmath$0.316 \pm 0.066$} \\
 & AbsErr$\downarrow$  & $1.891 \pm 0.133$ & \textbf{\boldmath$1.327 \pm 0.158$} \\
\midrule
\multirow{3}{*}{2}
 & Rec.$\uparrow$      & $0.124 \pm 0.085$ & \textbf{\boldmath$0.324 \pm 0.019$} \\
 & F1$\uparrow$        & $0.161 \pm 0.076$ & \textbf{\boldmath$0.335 \pm 0.017$} \\
 & AbsErr$\downarrow$  & $1.238 \pm 0.059$ & \textbf{\boldmath$0.958 \pm 0.018$} \\
\midrule
\multirow{3}{*}{3}
 & Rec.$\uparrow$      & \textbf{\boldmath$0.478 \pm 0.097$} & $0.437 \pm 0.012$ \\
 & F1$\uparrow$        & $0.389 \pm 0.033$ & \textbf{\boldmath$0.420 \pm 0.008$} \\
 & AbsErr$\downarrow$  & \textbf{\boldmath$0.617 \pm 0.106$} & $0.677 \pm 0.022$ \\
\midrule
\multirow{3}{*}{4}
 & Rec.$\uparrow$      & $0.438 \pm 0.051$ & \textbf{\boldmath$0.459 \pm 0.038$} \\
 & F1$\uparrow$        & $0.398 \pm 0.024$ & \textbf{\boldmath$0.444 \pm 0.026$} \\
 & AbsErr$\downarrow$  & \textbf{\boldmath$0.622 \pm 0.013$} & $0.643 \pm 0.041$ \\
\midrule
\multirow{3}{*}{5}
 & Rec.$\uparrow$      & $0.224 \pm 0.030$ & \textbf{\boldmath$0.402 \pm 0.016$} \\
 & F1$\uparrow$        & $0.278 \pm 0.017$ & \textbf{\boldmath$0.422 \pm 0.022$} \\
 & AbsErr$\downarrow$  & $1.167 \pm 0.063$ & \textbf{\boldmath$0.889 \pm 0.026$} \\
\bottomrule
\end{tabular}%
}
\end{table}

\subsubsection{Component ablations}
\label{subsec:ablation}
We conduct a sequential component ablation to validate the architectural choices underlying the proposed user-conditioned walkability perception framework. At each stage, all previously determined components are fixed at their best-performing configuration while a single component is varied, isolating the contribution of each design decision.

\textbf{Image Backbone.} We first evaluate four image backbones: CaFormer-B36~\cite{yu2023metaformer}, ConvNeXt-Tiny~\cite{liu2022convnet}, Swin-Tiny, and ResNet50~\cite{he2016deep}. These models span convolutional, hierarchical, and MetaFormer-based architectures, with the tabular encoder fixed at FT-Transformer, Transformer-based fusion, and loss at CORAL. As shown in Table ~\ref{tab:component_ablation}, Swin-Tiny achieves the highest mean validation QWK and the lowest MAE, while ResNet50 obtains comparable performance and slightly higher Within-1 accuracy. We select Swin-Tiny as the image encoder for the subsequent experiments based on its best overall validation performance.

\textbf{Tabular Encoder.} With Swin-Tiny fixed, we compare a standard MLP with FT-Transformer~\cite{gorishniy2021revisiting} as the respondent attribute encoder, while keeping the fusion module and loss function fixed as Transformer-based and CORAL, respectively. As shown in Table ~\ref{tab:component_ablation}, FT-Transformer outperforms the MLP encoder across all metrics, likely due to its per-feature tokenisation mechanism that projects each categorical respondent attribute into a dedicated embedding and learns pairwise inter-attribute interactions through self-attention, rather than relying on implicit feature interactions in a flat concatenated input~\cite{gorishniy2021revisiting}.

\textbf{Fusion Module.} With Swin-tiny and FT-Transformer fixed, we compare an MLP fusion head with a Transformer-based fusion module, while keeping the loss function fixed as CORAL. As shown in Table~\ref{tab:component_ablation},  the Transformer-based fusion head achieved superior performance.

\textbf{Loss Function.} Finally, with all architectural components fixed, we compare three loss functions: standard cross-entropy, CORN~\cite{shi2023deep}, and CORAL. CORN (Conditional Ordinal Regression for Neural Networks) models conditional probabilities, $(P(y > k \mid y > k-1))$, using independent output neurons and eligible training subsets for each ordinal threshold. As shown in Table~\ref{tab:component_ablation}, both ordinal loss functions outperformed cross-entropy, confirming that the five-point walkability scale contains meaningful rank-order information that is discarded by nominal classification losses. CORAL further outperformed CORN across all metrics. This result is notable because existing evaluations of CORAL and CORN have primarily focused on unimodal ordinal regression settings~\cite{CAO2020325,shi2023deep}, whereas the present study evaluates them within a multimodal image--tabular ordinal classification framework. CORAL was therefore selected as the loss function for all reported experiments.

\begin{table}[t]
\centering
\caption{Component ablation on the validation set. The selected model uses Swin-Tiny, FT-Transformer, fusion Transformer, and CORAL. W-1 denotes Within-1 accuracy.}
\label{tab:component_ablation}
\scriptsize
\setlength{\tabcolsep}{3.1 pt}
\renewcommand{\arraystretch}{1.30}

\begin{tabular}{@{}llcccc@{}}
\hline
Component & Variant & QWK & MAE & ACC & W-1 \\
\hline
Selected & Final 
& \textbf{0.490$\pm$0.015} 
& \textbf{0.763$\pm$0.021} 
& \textbf{0.421$\pm$0.018} 
& 0.843$\pm$0.004 \\
\hline
\multirow{3}{*}{Backbone} 
& ResNet50 
& 0.486$\pm$0.018 
& 0.774$\pm$0.003 
& 0.405$\pm$0.009 
& \textbf{0.851$\pm$0.005} \\
& ConvNeXt-T 
& 0.464$\pm$0.024 
& 0.812$\pm$0.038 
& 0.383$\pm$0.016 
& 0.836$\pm$0.022 \\
& CaFormer 
& 0.462$\pm$0.018 
& 0.790$\pm$0.027 
& 0.407$\pm$0.016 
& 0.836$\pm$0.008 \\
\hline
Tabular & MLP 
& 0.465$\pm$0.031 
& 0.793$\pm$0.029 
& 0.404$\pm$0.020 
& 0.838$\pm$0.010 \\
\hline
Fusion & MLP 
& 0.467$\pm$0.021 
& 0.777$\pm$0.024 
& 0.406$\pm$0.011 
& 0.843$\pm$0.009 \\
\hline
\multirow{2}{*}{Loss} 
& CORN 
& 0.444$\pm$0.011 
& 0.805$\pm$0.010 
& 0.405$\pm$0.005 
& 0.830$\pm$0.010 \\
& CE 
& 0.442$\pm$0.014 
& 0.828$\pm$0.016 
& 0.397$\pm$0.014 
& 0.818$\pm$0.006 \\
\hline
\end{tabular}
\end{table}

\begin{table}[!t]
\centering
\caption{Permutation-based importance of respondent-level features.}
\label{tab:perm_importance}
\scriptsize
\renewcommand{\arraystretch}{1.08}
\setlength{\tabcolsep}{2.2pt}
\resizebox{\columnwidth}{!}{%
\begin{tabular}{clccc}
\toprule
Rank & Feature & $\Delta$QWK$\uparrow$ &  $\Delta$MAE$\uparrow$ & $\Delta$W-1$\uparrow$ \\
\midrule
1 & Age & 0.204 (0.034) & 0.195 (0.032) & 0.074 (0.012) \\
2 & Residence type & 0.166 (0.020) & 0.154 (0.022) & 0.060 (0.008) \\
3 & Walking frequency & 0.164 (0.023) & 0.148 (0.023) & 0.062 (0.008) \\
4 & Childhood area & 0.125 (0.017) & 0.120 (0.020) & 0.047 (0.007) \\
5 & Gender & 0.114 (0.019) & 0.108 (0.017) & 0.041 (0.005) \\
6 & Disability & 0.072 (0.006) & 0.061 (0.006) & 0.025 (0.003) \\
7 & Childhood country & 0.067 (0.008) & 0.060 (0.010) & 0.026 (0.003) \\
\bottomrule
\end{tabular}%
}
\end{table}

\subsection{Attribute importance}
To examine the sensitivity of model performance to individual respondent attributes, permutation-based feature importance \cite{breiman2001random} was estimated by randomly shuffling each attribute's values in the test set while holding all others fixed. The resulting change in performance was measured using QWK, Within-1 accuracy, and MAE. For QWK and Within-1 accuracy, importance was defined as the decrease from the baseline score, whereas for MAE it was defined as the increase in prediction error. Each attribute was permuted 30 times per seed;  importance was first summarised as the median change across permutations for each seed, and then reported as the median with interquartile range (IQR) across all seeds. 

Table~\ref{tab:perm_importance} presents the results, which we interpret as a relative ranking of attribute influence rather than as independent, additive marginal contributions to model performance. Age produced the largest drop in QWK ($\Delta$QWK = 0.204, IQR = 0.034), followed by residence type and walking frequency with comparable importance scores 
($\Delta$QWK = 0.166 and 0.164, IQR = 0.020 and 0.023, 
respectively), indicating that lived environmental context and the 
habitual walking behaviour are both informative to the model's 
predictions. Notably, the $\Delta$QWK for age alone exceeds the total QWK improvement of the user-conditioned model over the image-only baseline (0.184; Table~\ref{tab:main_comparison}), which would not be expected if these values represented independent, additive marginal contributions. This is consistent with a documented limitation of permutation-based importance under correlated predictors, whereby shuffling one attribute independently of others (plausibly including age, residence type, childhood residential environment, and walking frequency) can generate unrealistic respondent profiles and inflate apparent importance beyond an attribute's true marginal contribution \cite{hooker2019please}. Accordingly, we do not interpret the magnitudes in Table~\ref{tab:perm_importance} as a decomposition of the total attribute contribution to model performance; instead, they should be viewed as indicating the relative influence of each attribute under this perturbation procedure.  Childhood area and gender showed moderate importance, while disability status and childhood country had the lowest importance scores, which may partly reflect their limited variability in the sample: 83.8\% of respondents reported no walking-related disability, and 80.0\% were raised in Australia. Overall, the relative ranking of attributes was broadly consistent across QWK, Within-1 accuracy, and MAE, supporting the robustness of the feature-importance pattern.

\section{Conclusion}
This paper introduced a new walkability perception dataset spanning urban, suburban, and regional areas in Australia, linking sidewalk-view imagery with individual rater attributes. A supplementary viewpoint comparison showed that sidewalk-view imagery received significantly higher walkability ratings than matched street-view imagery, indicating that the choice of visual data source is a substantive design decision in walkability perception surveys rather than an interchangeable convenience. The proposed multimodal user-conditioned framework, which conditions walkability perception predictions on both sidewalk-view imagery and rater characteristics, substantially outperformed an image-only baseline across all evaluation metrics. To the best of our knowledge, this is the first study to formulate visual walkability perception as a user-conditioned prediction task, demonstrating that respondent-level attributes provide a predictive signal beyond image content alone. Complementary interaction analysis showed that respondent attributes shape how individuals differentiate between environments rather than simply shifting overall rating levels, a pattern reinforced by permutation-based feature importance showing that predictive contribution is distributed across the respondent profile rather than driven by a single dominant attribute. These findings suggest that individual-level variation in perceived walkability is systematic and should not be treated merely as noise to be averaged out. Overall, the results support moving beyond aggregated, observer-independent walkability scores toward models that explicitly represent who is evaluating an environment, providing a step toward more inclusive approaches to walkability assessment.

This study has several limitations. The dataset is drawn from Australian locations and a predominantly Australian-raised respondent pool, and the transferability of both the perception patterns and the trained model to other cultural and infrastructural contexts remains to be established. The respondent attributes capture systematic, group-level sources of perceptual variation; fully distinctive preferences, which the modest inter-rater agreement suggests are substantial, are beyond the reach of our current dataset and study. The viewpoint comparison was a supplementary analysis intended to motivate the choice of sidewalk-view imagery rather than a full investigation of viewpoint effects; a systematic examination with broader imagery data and participant pools is left for future work.

\ifCLASSOPTIONcaptionsoff
  \newpage
\fi

\begin{IEEEbiographynophoto}{Moloud Damandeh}
received the B.Sc. and M.Sc. degrees in industrial engineering from Sharif University of Technology, Tehran, Iran. She is currently pursuing the
Ph.D. degree with the School of Civil and Environmental Engineering,
University of New South Wales (UNSW), Sydney, Australia, where she is a member of the Research Centre for Integrated Transport Innovation (rCITI).  Her research
focuses on measuring perceived walkability of urban
environments using pedestrian-perspective imagery, computer vision, and
multimodal deep learning.
\end{IEEEbiographynophoto}

\begin{IEEEbiographynophoto}{Meead Saberi}
 is a Professor in the School of Civil and Environmental Engineering at the University of New South Wales (UNSW), Sydney, Australia, a member of the Research Centre for Integrated Transport Innovation (rCITI), and an Australian Research Council Future Fellow (2026-2031). His work has advanced both theoretical and applied understanding of transport networks, urban mobility, and sustainable transport systems.
\end{IEEEbiographynophoto}

\end{document}